\documentclass[a4paper]{llncs}
\RequirePackage[l2tabu, orthodox]{nag}
\usepackage{subfig}
\usepackage{amsmath}
\usepackage{graphicx}
\usepackage[hyphens]{url}
\usepackage{fancyhdr}
\graphicspath{{img/}}
\begin{document}
\mainmatter{}  % start of an individual contribution

\title{The Artificial Mind's Eye}
\subtitle{Resisting Adversarials for Convolutional Neural Networks
  using Internal Projection}

\titlerunning{The Artificial Mind's Eye}

\author{Harm Berntsen\inst{1}%
%\thanks{Please note that the LNCS Editorial assumes that all authors have used
%the western naming convention, with given names preceding surnames. This determines
%the structure of the names in the running heads and the author index.}%
  \and Wouter Kuijper\inst{1}
  \and Tom Heskes\inst{2}}
\authorrunning{The Artificial Mind's Eye}

%\institute{Springer-Verlag, Computer Science Editorial,\\
%Tiergartenstr. 17, 69121 Heidelberg, Germany\\
%\url{http://www.springer.com/lncs}}

\institute{Nedap N.V. \and Radboud University}
\maketitle
\thispagestyle{fancy}

\begin{abstract}
We introduce a novel artificial neural network architecture that integrates
robustness to adversarial input in the network structure. The main idea of our
approach is to force the network to make predictions on what the given
instance of the class under consideration would look like and subsequently
test those predictions. By forcing the network to redraw the relevant parts
of the image and subsequently comparing this new image to the original, we
are having the network give a ``proof'' of the presence of the object.
\end{abstract}

\section{Introduction}

Convolutional Neural Networks (CNNs) have been shown to work well on image
classification tasks~\cite{ILSVRC15}. However, CNNs are vulnerable to adversarial
images~\cite{Nguyen2015,Szegedy2014}. In this paper we introduce a
novel type of network structure and training procedure that results in
classifiers that are provably, quantitatively more robust to adversarial
samples. Adversarial images can be found by perturbing a
normal image in such a subtle way that the change is usually
imperceptible by the naked eye~\cite{Goodfellow2015,Szegedy2014}.

The main idea of our approach is to force the network to make
predictions on what the given instance of the class under
consideration would look like and subsequently test those
predictions. Technically we achieve this by
chopping the classifier network into three stages: estimation, projection and
comparison.

The first stage estimates a vector of parameters (displacement,
rotation, scale and, possibly, various object specific internal
deformations) from the image. The second stage generates an image
based on the estimated parameters. The third stage compares the
projected image with the actual image and delivers a likelihood value
which can be turned into a verdict using a threshold. The working
hypothesis is that this network structure improves robustness against
adversarial samples.

There are two intuitions behind this working hypothesis. The first is
that parameter estimation is a \emph{smoother} task than
classification. Meaning that an orbit through the multidimensional
output space can be expected to have a smooth corresponding orbit
through the input space. In other words: it is possible to meaningfully
interpolate parameters \emph{for a model of a given class} but
it is much harder to meaningfully interpolate between \emph{models of
  two or more different classes}.

The second intuition behind our working hypothesis is that, by forcing
the network to draw a new image using only the estimated parameter
vector and subsequently comparing this new image to the original, we
are having the network give a ``proof'' of the presence of the
object. By carrying out this comparison only through myopic,
local features we ensure that, in order to get enough probability mass
to make the threshold, the network must be fairly precise in
reproducing the internal details of the objects. In effect we force
the network to learn much more than just a discerning set of features,
we force it to learn also the detailed internal structure of the
object, thereby making it inherently more robust against adversarial
input.

In this paper we lay the conceptual groundwork and give initial experimental
results. We hope that this will enable further research on combining our
approach with other, orthogonal, approaches like adversarial training and on
applying this method, or refinements inspired by it, on real- world tasks.

The source code for training the networks and to generate adversarial
images is available at
\url{https://github.com/hberntsen/resisting-adversarials}.

\subsection{Related Work}

Neural networks recognise objects in a different way than humans. As Ullman et
al.~\cite{Ullman2016} point out: ``\ldots the human recognition system uses
features and learning processes, which are critical for recognition, but are
not used by current models''. They show that where humans can recognise
internal components of the objects in the image, current neural networks do
not. With knowledge about the internal representation of the objects, false
detections can be rejected when it is not consistent with the internal
representation of the object. This corresponds with the sensitivity to
adversarial images with an imperceptible change that have been shown
in~\cite{Szegedy2014} and various work since~\cite{Nguyen2015}. They show that
the smoothness assumption does not hold for neural networks; an imperceptible
change in the query image can flip the classification. Goodfellow et al.\ argue
that the primary cause for this is the linear behaviour of the networks in
high-dimensional spaces~\cite{Goodfellow2015} as opposite to the nonlinearity
suspected in~\cite{Szegedy2014}. The adversarial images are not
isolated, spurious points in the pixel space but appear in large regions of the
space~\cite{Tabacof2015}. Moreover, adversarial images can be efficiently
computed using gradient ascent, starting from any input~\cite{Goodfellow2015}.

Though the existence of adversarial examples is universal~\cite{Szegedy2014},
neural networks can be made more robust against them. One way is to include
adversarial examples in the training
data~\cite{Goodfellow2015,Huang2015,LYU2015,Szegedy2014}, e.g.\ by assigning
them to an additional rubbish class. Apart from increasing the robustness it
can also increase the accuracy on non-adversarial examples. Another approach is
to adapt the model of the network to improve
robustness~\cite{Chalupka2015VisualCF,Gu2014TowardsDN}.
In~\cite{Chalupka2015VisualCF} the authors identify features that are causally
related with the classes. Their learning procedure could be seen as a way to
train a classifier that is robust against adversarial examples. 
In~\cite{Gu2014TowardsDN} the authors test several denoising architectures to
reduce the effects of adversarial examples. They conclude that the sensitivity
is more related to the training procedure and objective function than to model
topology and present a new training procedure.

We use 3D models to train the classifier. Though this is artificial data, it
can be used as training material for real data, e.g.\ for object
detection~\cite{Peng2015,Sun2014} or even aligning 3D models within an 2D
image~\cite{Aubry2014,Massa2015,Su2015}. The work of~\cite{Aubry2014} does this
using HOG descriptors, while~\cite{Massa2015,Su2015} use neural networks.
They have trained a CNN to predict the viewpoint of 3D
models and were successful in applying this model to real-world images.

\section{Network Architectures}
In this section we describe the network architectures that we use to test the
robustness of our approach. The task of each network is the same: classify the
image. Our data consists of greyscale ImageNet images where a part of the image
is overlaid with an alpha-blended instance of a 3D model. We use three 3D
models that are parametrised by their Euler rotation. The neural network has to
recognise the 3D models in all those rotations and emit which 3D model, if any,
is visible in the query image. We compare the robustness against adversarial
images using three concrete network structures.  We will refer to the three 3D
models as positive classes and refer to the `None' class as the negative class.

\subsection{Networks}

\subsubsection{Direct Classification}
To set a baseline, we train a network to map the query images
directly to a probability distribution over the classes. This network is based
on AlexNet~\cite{Krizhevsky2012}, which has been shown to work well in various
situations~\cite{Norouzi2014,Simonyan2013,Su2015,Tang2015}.
%We reduced the size of layers to adapt the network this network to our use
%case.
To adapt AlexNet to a reduced set of classes and smaller query image, we use a
reduced version of AlexNet from~\cite{Berntsen2015} which uses smaller layers.
We replaced the last layer with a softmax classification layer. The softmax
layer has four outputs, three for the positive classes and one to indicate the
negative class.

\subsubsection{Direct Classification + Parameter Estimation}
The Direct Classification + Parameter Estimation network is a variant
of the Direct Classification network that has an additional
output: the parameters of the model. This additional output forces the neural network to develop a better understanding of
the 3D models it has to recognise.  The parameter estimation is only used to guide the
training process and is not used after the network has been trained.

\subsubsection{Triple-staged}

\begin{figure}
\centering{
    \includegraphics[width=\linewidth]{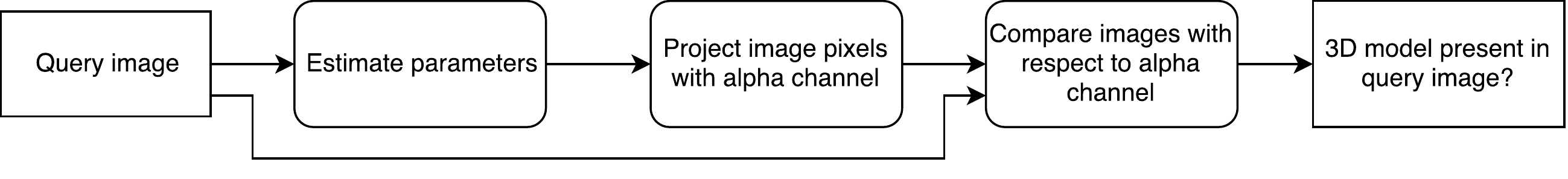}
    \caption{Data flow diagram of the triple-staged network structure
      for a single 3D model. The model parameters are estimated from the input
      image, converted back to an image and then compared to the original
      image.}\label{fig:architecture} }
\end{figure}

We will first describe the triple-staged network as if it is specific to one
single class. We expand this design later to a configuration for multiple
classes. The triple-staged network contains three stages: (1) estimation, (2)
projection, (3) comparison, shown in Fig.~\ref{fig:architecture}. Each stage
was trained separately and finally merged into one network.

The first stage maps the query image to a parameter vector that describes a 3D
model. In our running example the parameters describe just the Euler
rotation of a 3D model but in general this can also include scale,
pan, and internal parameters such as dimensions, and rotational and
linear joints. The estimations are clipped to their valid range. The
network structure of this stage is the same as the direct
classification + parameter estimation network without the task to
predict the class.

The second stage of the network projects the parameter vector to a 2D image
that
contains the rendered 3D model in front of a black background. The
alpha channel indicates to which degree each pixel belongs to the 3D
model. In~\cite{Dosovitskiy2015}, it was shown that a deep,
deconvolutional neural network can be trained to generate images that
are parametrised by a broad set of classes and viewpoints. Due to our
smaller set of classes and parameters, we use a downscaled variant of
the 1s-S network from~\cite{Dosovitskiy2015}. The first and second stage together form an
autoencoder where the bottleneck contains an understandable instantiation
vector of the input. This concept was already applied in the context of
transforming autoencoders~\cite{Hinton2011transforming}.

\begin{figure}[h]
    \begin{tabular}{cc}
    \includegraphics[scale=0.5]{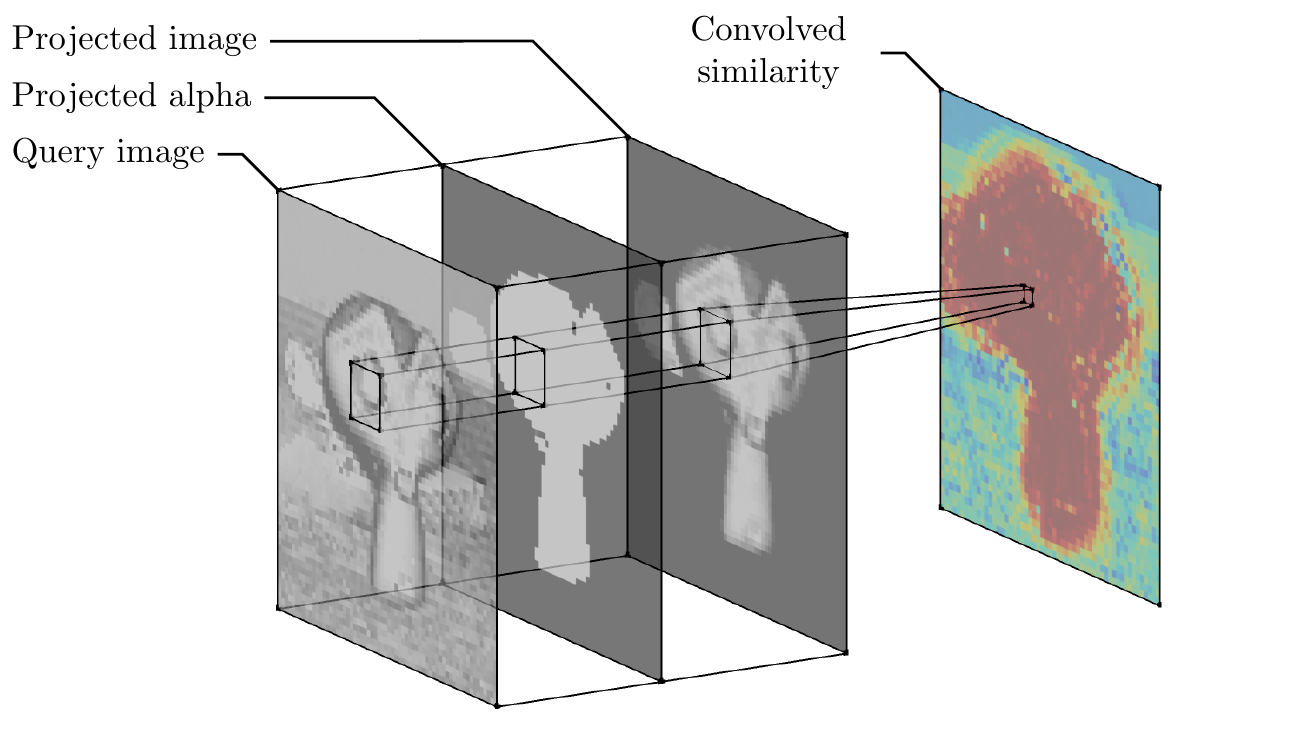} &
    \includegraphics[scale=0.5]{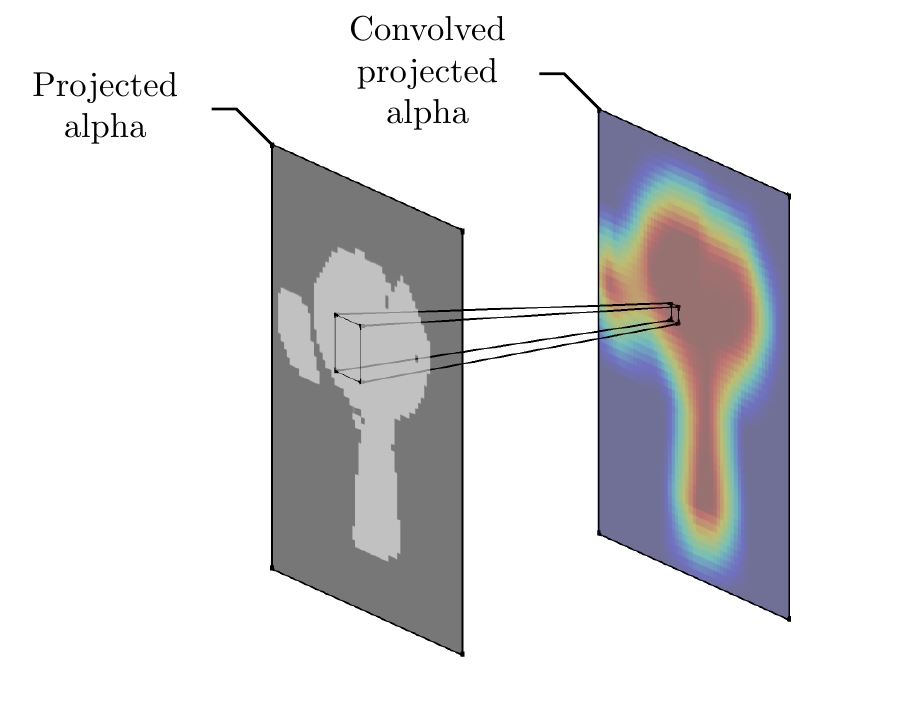} \\    
    \includegraphics[scale=0.5]{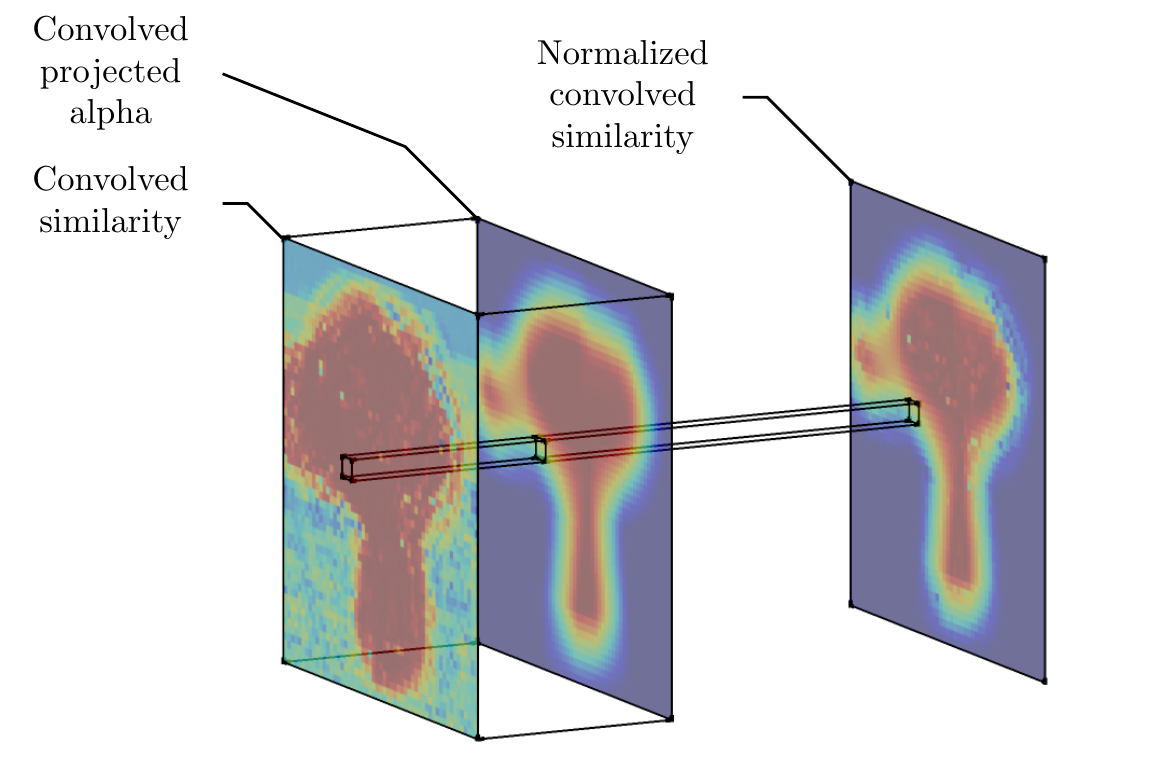} &
    \includegraphics[scale=0.45]{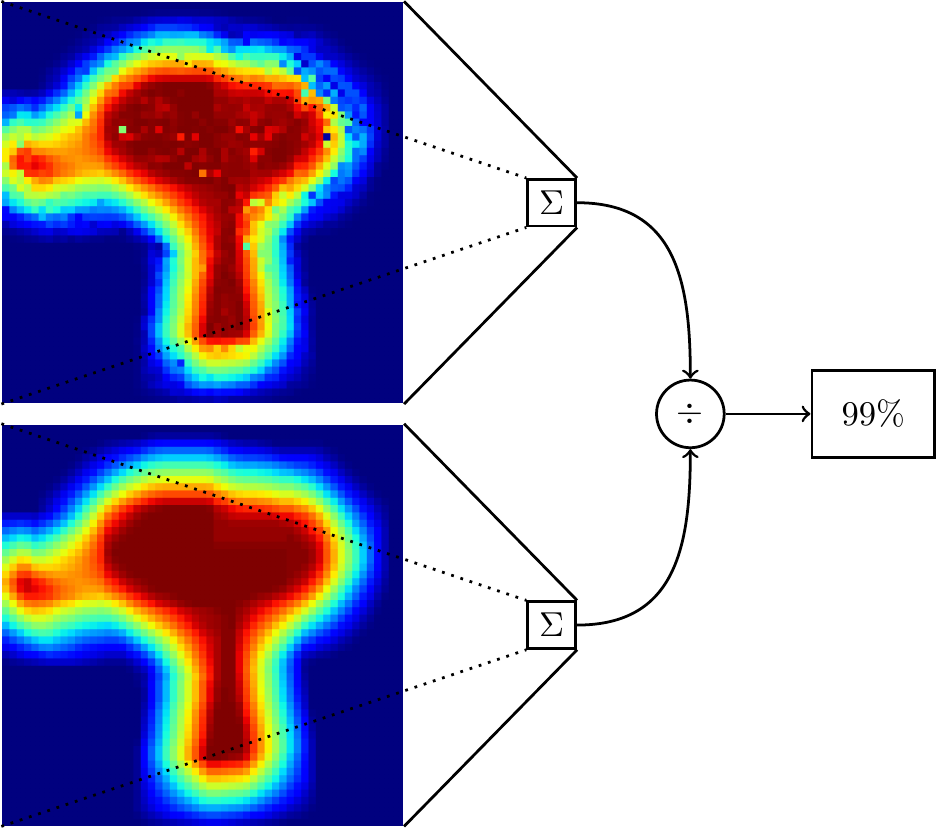}
    \end{tabular}
    \caption{Visualisation of the comparison stage of the network. We
      convolute this stage over every $10 \times 10$ pixels patch with
      stride 1. This generates the \emph{convoluted similarity
        map}. Next to this similarity map, we apply a $10 \times 10$
      average pooling layer over the projected alpha channel to
      generate the \emph{convoluted, projected alpha channel}. We can
      then directly multiply the convoluted alpha channel and the
      convoluted similarity map to end up with the \emph{normalized,
        convoluted similarity map.} Next we feed both the normalized,
      convoluted similarity map and the convoluted, projected alpha
      channel, as a whole, into a single-output, sum--reduction layer.
      Finally we obtain the ratio by having a single output multiply
      the similarity sum with the reciprocal of the total weight
      sum. To obtain a final verdict we apply a threshold $\Theta$ over the
      output (cf. Section~\ref{sec:results})}\label{fig:architecture_stage3}
\end{figure}

The final stage has to compare each
projected image with the query image. Here we
follow~\cite{Zagoruyko2015}, which shows how to compare image patches
using CNNs. We adapted the 2-channel
structure to create a network that compares $10 \times 10$ pixel image
patches with respect to the alpha channel. This network is
convoluted over the output of the second stage, giving it only
local data to work with. The network was trained to emit a binary
output that indicates whether the original and projected image patch
should be considered equal. Fig.~\ref{fig:architecture_stage3} visualises how
this network works.

The combination of the three stages is capable of determining the presence of a
single class in the query image. This does not scale well since a separate
network has to be trained for each and every class. This issue is addressed by
adding the class to generate as a parameter to the projection stage of the
network. Each stage can now be trained on data of all the 3D models at once.
The class parameter improves scalability of the triple-staged network since
only one network needs to be trained for multiple classes. To generate a
classification for a query image, it is provided as an input to the network
multiple times with a different class parameter. If there is any class where the
output of the network rises above the threshold $\Theta$, we use the class that
yields the highest similarity score. Otherwise we judge that none of the
classes are visible in the query image.

\begin{figure}
\centering{
    \includegraphics[width=\linewidth]{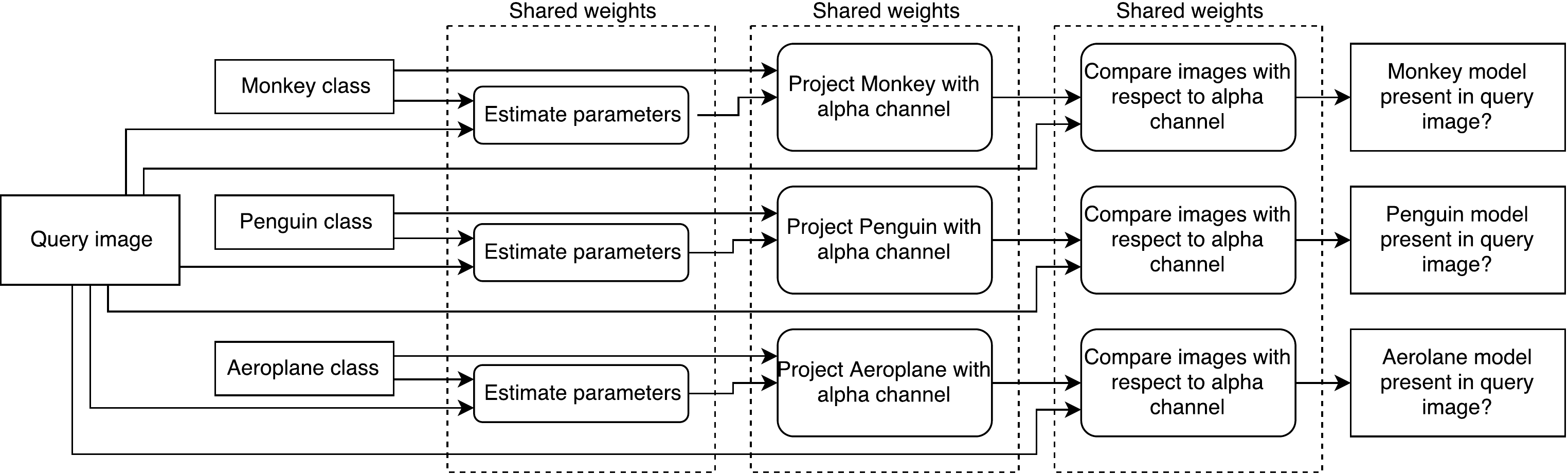}
    \caption{Architecture of our triple-staged network.
    Each class shares the weights in the network but hardcodes the class
    input parameter of the network.}\label{fig:architecture_full}
}
\end{figure}

Similarly, an optimisation we applied is to supply the class parameter to the
prediction stage of the network. We add the class as additional binary
channels to the query image. Without the class information passed to
this network, the network would internally need to determine which
class is visible in the query image. We found that supplying the class
information to the network increased the robustness of the
triple-staged network. Figure~\ref{fig:architecture_full} shows the
triple-staged network architecture we used.

\subsection{Rationale}

The main feature of the triple-staged architecture is to be robust against
adversarial samples that cause the network to indicate that a certain class is
visible when it is not. To let the neural network produce a false positive classification, an
adversary needs to perturb the image such that it ultimately fools the
comparator stage of the network. However, in order to do so, it must
pass through both the estimator and the projector stage.
Any attempt to generate a false positive will start
drawing another 3D model over the existing one because the comparator
compares the query image with the stable internal projection of the
class in question. Ultimately then, the `false positive' class will be
evidently visible in the query image.

Since the comparator network directly consumes the query image, this
network could still be susceptible to adversarial perturbations of the
query image in much the same way as a normal classifier would be. In
order to reduce susceptibility, we limited the input space of
the network to a single $10\times10$ pixel patch. This is enough to
learn the general concept of two patches being ``similar'' (modulo
some minor deviations and/or artefacts) but it is not enough to learn
longer range correlations in the query image (that would give the
adversarial a clear gradient to follow in generating adversarial
input). We convolve this local network across the whole image, hence
an adversary would need to simultaneously fool sufficiently many
individual, local patch comparisons to make a significant impact on
the overall similarity mass.

\section{Experiment Set-up}

In this section we describe how we will test the network architectures from the
previous section for their robustness.

\subsection{Training method}

As objects to recognise, we use parametrised 3D models. We rendered
$64 \times 64$ pixel greyscale
images of three 3D models: a Monkey (the Suzanne model
from~\cite{Blender}), Penguin~\cite{PenguinModel} and an
Aeroplane~\cite{AeroplaneModel} using Blender~\cite{Blender}. 
We took the rotation of the 3D model over three axes as our
parameter space though our method is not limited to this. The
rotations were uniformly sampled from the range of $[-0.5, 0.5]$ radians. To give
the 3D models a `natural' background, we use alpha-composition to
blend the 3D model in front of randomly sampled images from the
ImageNet dataset~\cite{ILSVRC15}.  This reduces overfitting of the
network on the otherwise black background. We generated 
$4 \times 10^4$ samples for each class. The
None class simply consists of random ImageNet images.

We used Caffe~\cite{Jia2014} for the network implementations. The direct
classification networks were trained using all $16 \times 10^4$ samples.
We left the predicted parameter vector for the negative samples undefined.
Each stage of the triple-staged was trained separately. The estimator stage was
only trained on the subset of positive samples. The second stage
was trained on the original data as rendered through Blender. The input of
this stage consists of the binary encoding of the class and the rotation
parameters.

The data for the third stage of the network was generated by passing data 
through the first two stages of the network. This resulted in a new dataset
with the query image, ground truth class, projected image and projected alpha
mask. From this data we generated a balanced dataset where half of the samples
should be considered the same and the other half of the samples is not.
The samples that are considered different compare the query image, which is
either a random ImageNet image or one of the 3D models in front of an ImageNet
background, against one of the projected images by the second stage of the
network. From this training set we sampled $10 \times 10$ pixel patches where
the projected alpha mask indicated that at least $1\%$ of the pixels belonged
to the model. The other samples do not matter since their comparison is
cancelled out by the multiplication with the projected alpha (visualised in
Fig.~\ref{fig:architecture_stage3}).

\subsection{Adversarial Image Generation}\label{sec:adversarial_input}
When we want to generate an adversarial query image $\tilde{x}$, we
search for a minimal perturbation of the original image $x$ that is
sufficient to flip the classifier towards a chosen adversarial target
class value $y$. To do this we adopt the fast gradient sign method
of~\cite{Goodfellow2015}. The fast gradient sign method can
efficiently generate adversarial images using backpropagation.  Our
aim is to generate adversarial samples that flip the classification to
another positive class value $y$. Specifically, we perturb an image by
computing
\[
    \tilde{x} = \text{clip}(x - \text{sign}({\nabla}_x J(\theta, x, y)), 0, 255),
\]
with $J$ the loss function over the query image $x$ and network
parameters $\theta$. Since we use 8-bit
greyscale images with a range of $[0,255]$, each pixel of the image will be
minimally perturbed. This function is applied as often as needed to flip the
classification of the network to the target $y$.

\section{Results}\label{sec:results}

We test our networks on a separate test set which consists of
10000 samples of each class. The backgrounds are sourced from the ImageNet
validation set.
%We first test the error rate of the networks in normal
%circumstances. We then continue with testing adversarial input.
We first measure the classification performance of the networks on
non-adversarial images, see Table~\ref{tab_classifier_error_rate} for
the results. The direct classification networks have the lowest error
rate, followed by the triple-staged network. With $\Theta = 0.2$ or
$\Theta =0.7$, the `None' class is chosen more/less often. Although
choosing $\Theta$ somewhere in the middle seems to be the best option
purely in terms of minimizing classification error, our data clearly
shows that there is a trade-off to consider concerning robustness to
adversarial samples versus classification error.

\begin{table}
    \centering
    \caption{Classification error rate of the networks.}\label{tab_classifier_error_rate}
    \begin{tabular}{ll}
        \hline\noalign{\smallskip}
        Network                          & Error rate \\
        \noalign{\smallskip}
        \hline
        \noalign{\smallskip}
        Direct classification            & $0.01\%$\\
        Direct classification + Parameter Estimation     & $0.01\%$\\
        Triple-staged, $\Theta = 0.2$    & $1.34\%$\\
        Triple-staged, $\Theta = 0.45$   & $0.57\%$\\
        Triple-staged, $\Theta = 0.7$    & $3.09\%$\\
        \hline
    \end{tabular}
\end{table}

To compare the networks under adversarial conditions, we measure how many
iterations of adversarial perturbation it takes to change the classification
and how much
the image was changed. Here we follow~\cite{Szegedy2014} who measure
the amount of perturbation in adversarial sample for original sample
as distortion which is defined as:
\[
    \frac{1}{255}\sqrt{\frac{\sum_i{(\tilde{x}_i - x_i)}^2}{n}},
\]
where $x$ is the original image, $\tilde{x}$ is the distorted image
and $n$ is the number of pixels.

\begin{figure}[h]
    \includegraphics[width=\textwidth]{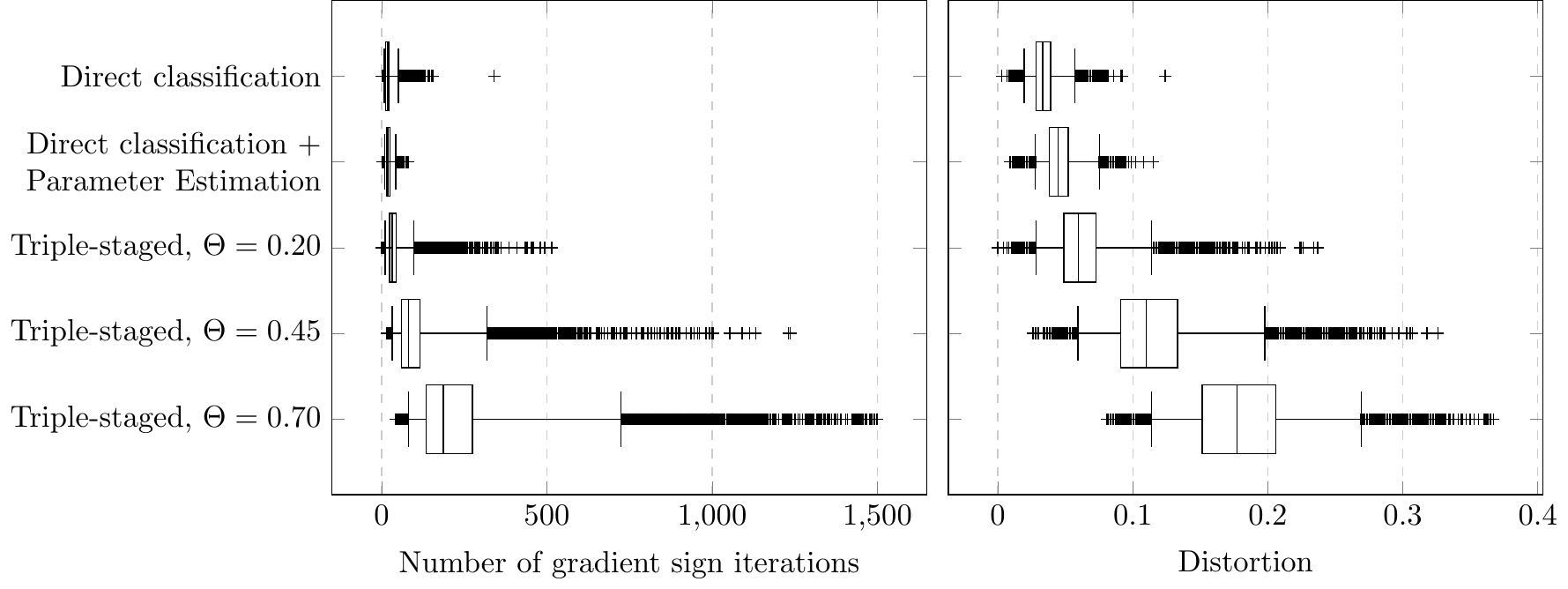}
    \caption{The effort required to convert the test images that
      contain one of the 3D models to an image where the network
      judges that another class of the 3D models is visible. There
      were no cases with 0 steps. For readability we clipped the number of
      steps in the graph to a maximum of 1500. The whiskers are
      placed at the 2\textsuperscript{nd} and 98\textsuperscript{th}
      percentile.}\label{fig:notnone_to_notnone}
\end{figure}

\begin{figure}
    \centering
    \subfloat[][Direct classification + parameter estimation]{
        \includegraphics[width=\linewidth]{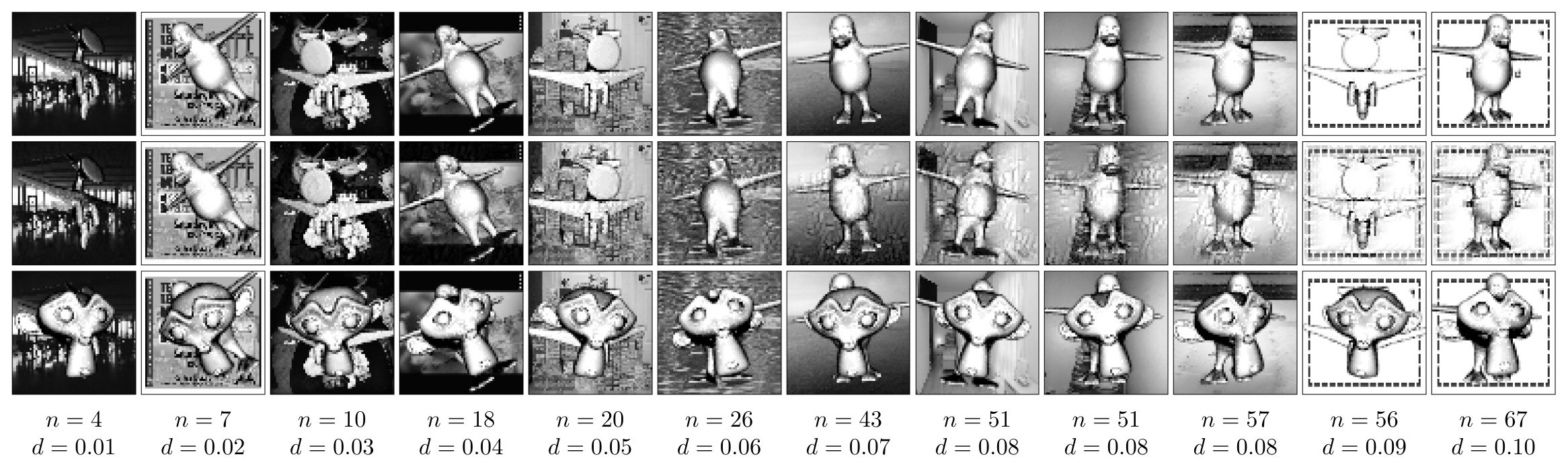}
        \label{fig:examples_directclassificationp}
    }
    \qquad
    \subfloat[][Triple-staged network with $\Theta = 0.7$]{
        \includegraphics[width=\linewidth]{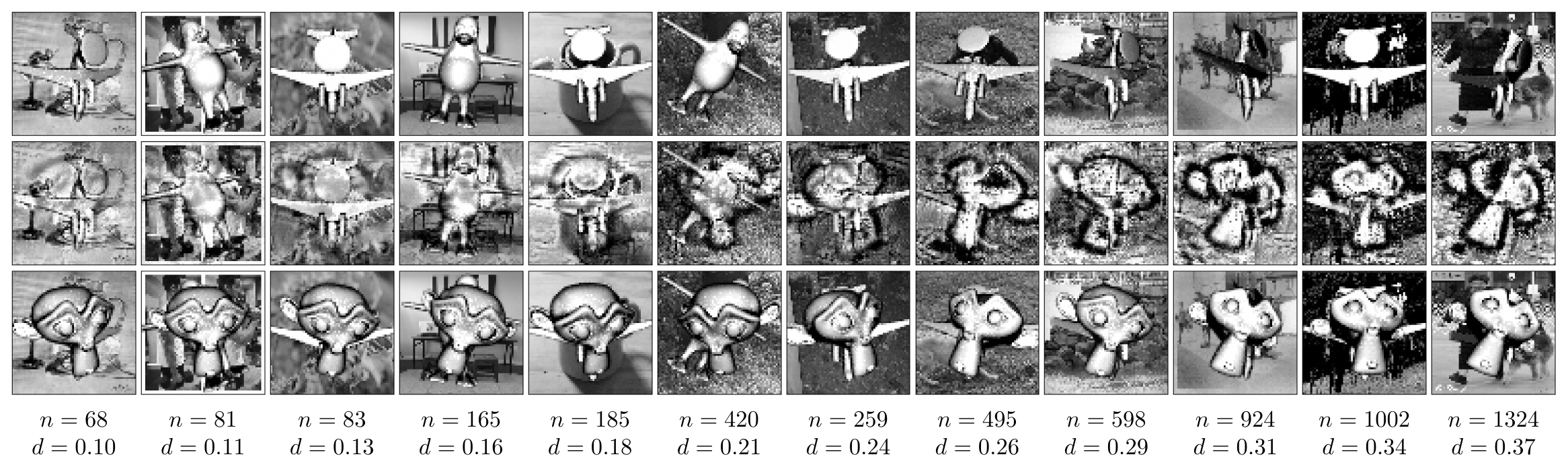}
        \label{fig:examples}
    }
    \caption{Generated adversarial images, ordered by distortion. The top row
    contains the original image, the second row the adversarial variant. The
    bottom row contains the original image with the Monkey 3D model
    alpha-blended, rendered by Blender using the predicted parameters. Every
    column is annotated with the distortion $d$ and number of iterations $n$.
    All adversarial samples are classified as the Monkey class.}
\end{figure}

We performed experiments with both false positive and false negative
adversarial images. To generate a false positive adversarial image, we start
from a test image containing one of our 3D objects, say a Monkey. Following the
procedure explained in Section~\ref{sec:adversarial_input}, we then construct
an adversarial image that makes
the network believe that the image belongs to the other class, once for a
Penguin and once for an Aeroplane. We repeat this procedure for all test images.
Figure~\ref{fig:notnone_to_notnone} shows the results for the false positive
adversarial images. The figure shows that for the direct classification
networks the required changes are limited:
the median of their distortion is still below
$0.1$ which indicates that the adversarial image is still very similar to the
original one. The examples in Fig.~\ref{fig:examples_directclassificationp} show
this. In contrast, the triple-staged network
requires significantly more effort to change the classification. The higher the
threshold, the more the query image needs to look like the internally projected
image. Figure~\ref{fig:examples} shows false positive adversarial samples for the $\Theta =
0.70$ network. The triple-staged network structure requires the adversary to
generate images that really start to look like the adversary class. As
Table~\ref{tab_classifier_error_rate} shows, the error rate on normal samples
is still reasonable at this threshold.

\begin{figure}
    \centering
    \includegraphics[width=\textwidth]{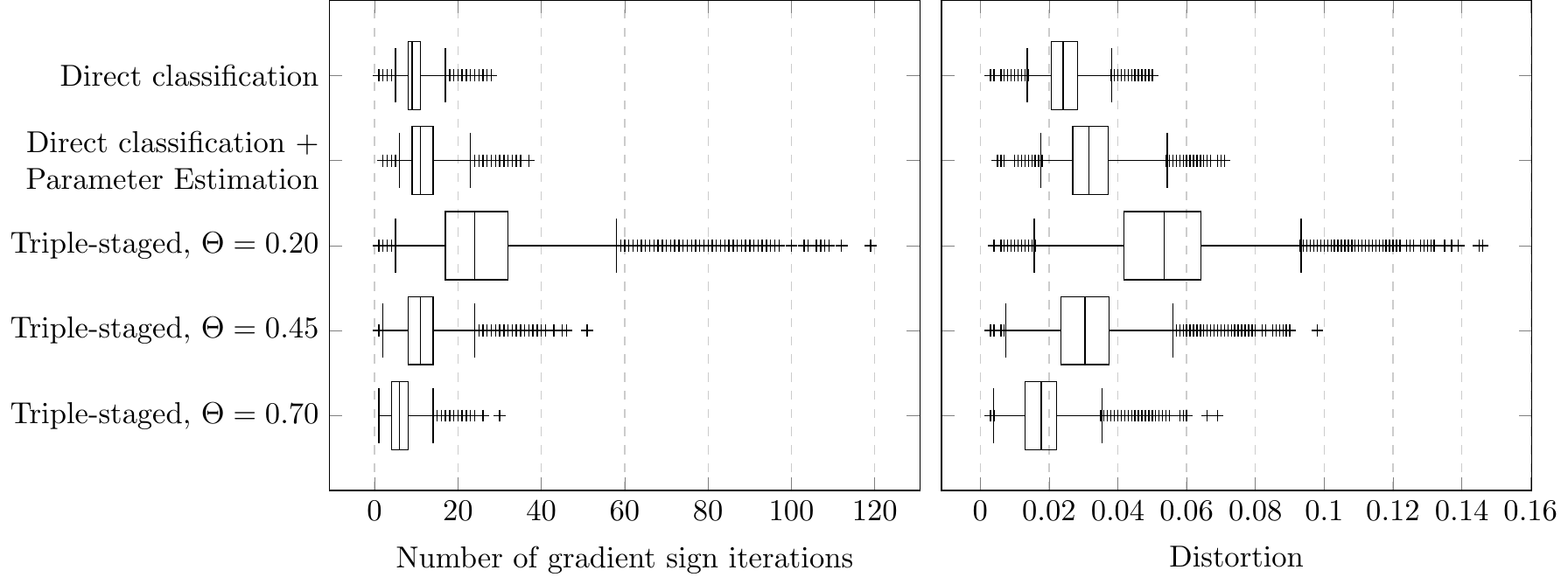}
    \caption{The effort required to convert the test images that contain
        one of the 3D models to an image where the network judges that none
        of the 3D models is visible. This was possible for all the samples in
        the test set. The instances that were misclassified in the first place
        were filtered out. This is
        $0.00\%, 0.01\%, 0.17\%, 0.77\%, 4.12\%$ of the data respectively from
        top to bottom. The whiskers are placed at the 2\textsuperscript{nd} and
        98\textsuperscript{th} percentile.}\label{fig:tonone}
\end{figure}

When we generate false positives, we start with an image that contains one of
the classes and perturb it to an image that is classified as `None'.
For the direct classification networks this is the 4\textsuperscript{th} class
they can predict. In the case of the triple-staged network, the output for every
class has to be below the threshold $\Theta$. Figure~\ref{fig:tonone} shows that the
number of required iterations is significantly higher for $\Theta = 0.20$
compared to the other networks.  Note that in contrast to the false positives,
the false negative adversarial samples are better resisted using a lower
threshold.  By lowering the threshold, the triple-staged network is less likely
to switch to the `None' class, requiring more work from the adversary.

%Though the clipping layer prevents out-of-range values, the adversary can still
%change the predicted
%parameters to a different but valid parameter vector. This allows the adversary
%to efficiently change the classifier's output from one of the classes
%corresponding to the 3D model to `None'.  The second stage projects an image
%which is based on incorrect parameters and thus we fail in the third stage
%because the original image and projected image do not match. We explicitly did
%not let the first stage predict the class as the results show that a direct
%classifier is relatively easy to fool. The higher threshold for the network
%output allows us to accept a larger change in the rotation of the model,
%requiring the adversary to introduce significantly higher changes in the image
%rotation. We believe that the smooth task of the parameter prediction improves
%robustness here.

\section{Discussion}
%kijken naar predictor, hoeveel %adversarials is uit de range?

We have adapted the classical network structure for classifier tasks
and shown significant improvements in robustness against adversarial
samples. In order not to pollute the results we have not taken into
account other types of solutions against adversarial samples such as
adversarial training. This does not however mean these techniques
would not be useful also in our setting. As future work we therefore
plan to incorporate adversarial training into our approach.

Future work could also apply our technique to include more parameters
including internal deformations, using joints etc. We have only tested
three 3D models with a limited parameter space.
%Further research is
%needed to check that our solution can indeed scale to more realistic settings.
In~\cite{Massa2015,Su2015}
it was already shown that it is possible to estimate viewpoints of 3D models in
real-world images. Dosovitskiy et al.~\cite{Dosovitskiy2015} have shown that a
deconvolutional neural can generate images based on many classes and
viewpoints. This opens up possibilities to expand our work to a real-world
situation.

For the present work we opted to train the network in three separate
stages. This allowed us quite a bit of control over the network
architecture which, in turn, allowed us a shorter route to testing the
working hypothesis. Nevertheless, as future work, it would be
interesting to develop end-to-end training methods for which the
architecture would be more emergent and less explicit. As an obvious
first step we could conceive of training the first two stages
end-to-end, as an autoencoder, instead of using manually parameterized
3D models.

%\begin{itemize}
%    \item The second stage of the network was trained on samples rendered by
%        Blender to get the best mapping from class to image. An alternative
%        approach could have used the prediction of the first stage of the
%        neural network, allowing the second stage to adapt itself to the small
%        mistakes the first network has made in the parameter prediction.
%        \todo{hoe kan ik deze tekst het best verwerken?}
%    \item Use adversarial samples during training, voor robustness maar ook
%        voorkomen noodzaak van imagenet zoals in mijn thesis.
%    \item Iets met clipping tussen fase 1 en 2 voor `weet niet' klasse?
%\end{itemize}

\section*{Acknowledgements}
We would like to thank Daan van Beek for the lively discussions during
the preconceptual stage of this work, and also for his detailed
comments on a draft of this paper.

\bibliographystyle{splncs03}
\bibliography{library}

\end{document}